\documentclass[11pt,bibtotoc]{scrartcl}

\usepackage[utf8]{inputenc}
\usepackage[T1]{fontenc}
\usepackage{lmodern}
\usepackage{a4wide}
\usepackage{hyperref}
\usepackage{graphicx}
\usepackage{subfigure}
\usepackage{listings}
\usepackage{amsmath}
\usepackage{marvosym}
\usepackage{textcomp}
\usepackage{float}
\usepackage[usenames,dvipsnames]{color}
\usepackage{multirow}
\usepackage{rotating}

\title{Exploring Twitter Hashtags}
\author{Jan Pöschko\\
\href{mailto:poschko@kth.se}{poschko@kth.se}
}

\newfloat{listing}{t}{listings}
\floatname{listing}{Listing}

\lstdefinelanguage{regexp}{
basicstyle=\ttfamily\small,
morecomment=[s]{(?\#}{)},
commentstyle=\color{NavyBlue},
} 

\newcommand{\h}[1]{\texttt{\##1}}

\begin{document}

\maketitle

\abstract{
Twitter messages often contain so-called \emph{hashtags} to denote keywords related to them. Using a dataset of 29 million messages, I explore relations among these hashtags with respect to co-occurrences. Furthermore, I present an attempt to classify hashtags into five intuitive classes, using a machine-learning approach. The overall outcome is an interactive Web application to explore Twitter hashtags.
}

\tableofcontents

\section{Introduction}

\emph{Twitter}\footnote{\url{http://twitter.com/}} is a fast-growing social Web application allowing its users to publish and communicate with very short messages, so-called \emph{tweets}, limited to 140 characters each. In the first half of 2010, there were over 100 million users registered at Twitter \cite{economictimes}, composing more than 65 million tweets per day \cite{twitterstat}.

Naturally, the language used in tweets is characterized by many abbreviations (e.g. \verb!4 U!) and emoticons (e.g. \verb!:)!), like in SMS. However, there are also very Twitter-specific forms of annotations, most notably so-called \emph{@-replies} and \emph{hashtags}, like in the following tweet:

\begin{quote}
\begin{verbatim}
@merazindagi Thanks! Will make more 4 U. Live performances in
#boulder area will be on http://saxy.us :) #jazz #rock #funk
#dance #livemusic
\end{verbatim}
\end{quote}

Hashtags are simply words that are preceded by a hash (\verb!#!). They can be used both inside the text and at its end to annotate keywords for a tweet. Twitter displays each hashtag as a link to a page listing other tweets containing the hashtag; that is where the ``tag'' in ``hashtags'' comes from, as they serve a similar purpose as tags on websites like Flickr\footnote{\url{http://www.flickr.com/photos/tags/}} and Delicious\footnote{\url{http://www.delicious.com/?view=tags}}.

The problem with many hashtags is that, just from their name, it is often impossible to tell what they are about (e.g. \h{tcot}, \h{p2}, \h{sgp}). This problem might (at least partially) be solved by the two approaches described in this work: a dictionary built upon co-occurrences (section \ref{cooccurrences}) and a machine-learnt classification into basic classes (section \ref{classification}), plugged into an interactive Web application (section \ref{webapp}).

\section{Dataset and pre-processing}

\begin{figure}
\centering
\includegraphics[width=\textwidth]{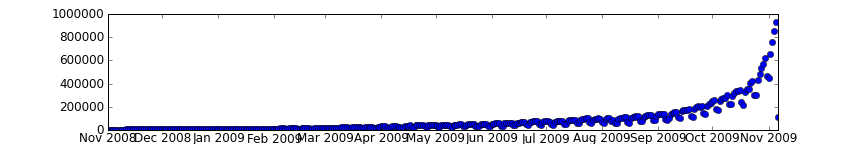}
\caption{Timeline of tweets per day in the provided dataset.}
\label{timeline}
\end{figure}

A nice feature of Twitter, at least from a researcher's point of view, is its open API to access tweets. It just can take a long time to crawl many tweets due to rate limits. I was lucky to get a dataset of 29 million tweets from Munmun De Cloudhury \cite{cloudhury}, crawled from November 2008 to November 2009, with the majority of tweets crawled in the last month, as can be seen in Figure \ref{timeline}. Although this bias towards later tweets is unlikely to have effects on the studies of hashtag co-occurrences and classification, this could be subject to further investigations. 

To reduce noise, I focused on those 85,503 hashtags (out of about 310,000) 
that occur in at least three tweets in this dataset. They correspond to 2,800,027 tweets where at least one of these ``relevant'' hashtags occurs.


All programming was done in the \emph{Python} programming language\footnote{\url{http://www.python.org/}}, which is particularly attractive for language processing because of the \emph{Natural Language Toolkit} (\emph{nltk}) \cite{nltk}. To handle the large dataset, a \emph{Whoosh}\footnote{\url{http://bitbucket.org/mchaput/whoosh/}} index of all the relevant tweets was built, esp. to have fast access to the tweets containing a certain hashtag.

As the language used in tweets differs in some ways very much from ordinary languages, a special way of tokenizing texts was needed. Especially, hashtags, @-replies, and URLs should be preserved in the tokenization process. See Listing \ref{wordre} for the custom regular expression that was used to find all the tokens in a given text. This differs from the maybe more common approach of splitting the text at certain delimeters, which I found impractical as e.g. a slash (\verb!/!) might denote a delimeter (as in \verb!this/that!) as well as an abbreviation marker (\verb!b/c!), or it might be used inside links.

\begin{listing}
\begin{lstlisting}[language=regexp,escapechar=^,frame=tb,tabsize=8,columns=fixed]
c/o|b/c|w/o|w/|\+/-|	(?# common abbreviations, +/- )
\d+(?:[.,:/-]\d+)+|	(?# numbers, fractions, dates, etc. )
(?:[:;][-=]?|=)[Dp(|)][D()]*|<3+|	(?# smileys )
(?:https?\:\/\/|www\.)[a-zA-Z0-9/.?=&\-#]*[a-zA-Z0-9/]|
			(?# URLs )
[#@]\w+|		(?# hashtags, @-replies )
\w+(?:-\w+)*(?:'\w+)?|  (?# ordinary words )
[$^\textsterling\EUR\textyen\textcent\textsection^@&%\+~]+         (?# special symbols )
\end{lstlisting}
\caption{The regular expression used to find words in a tweet text. Spaces, linebreaks, and comments were added for readability; the standard Python regular expression syntax is used.
}
\label{wordre}
\end{listing}

\section{Hashtag co-occurrences}
\label{cooccurrences}

\subsection{Dictionary}

Inspired by the \emph{Web 2.0 dictionary} \cite{web20dict}, I built a ``dictionary'' of hashtags defined by co-occurring hashtags. The co-occurrence count of two hashtags $h_i$, $h_j$ can be formally defined as
\[ C(h_i,h_j):=\left|\{\text{tweet }t\mid h_i\in t\wedge h_j\in t\}\right|.\]
Given a hashtag $h$, let its dictionary entry $D(h)$ consist of the ten hashtags $h_j\neq h$ with the highest co-occurrence counts $C(h,h_j)$, in descending order.

In total, there are 1,462,215 pairs of hashtags which co-occur. Storing their co-occurrence counts $C(h_i,h_j)$ in a Python dictionary consumes about 190 MB of memory, which would be feasible for single calculations, but probably not in a Web application, as introduced in section \ref{webapp}, running on a (multi-site) webserver with very limited resources. Apart from that, a memory-based storage does not scale well, and thinking of this application as a prototype for a more extensive version employing more data, another solution had to be found. That is why I created an SQLite\footnote{\url{http://www.sqlite.org/}} database file for storing the co-occurrence counts of all relevant hashtags.

\subsection{Evaluation}

\begin{table}
\centering
\subfigure[\h{obama}]{
\begin{tabular}{lr}
$h_j$&$C(h,h_j)$\\
\hline
\h{tcot}&1634\\
\h{teaparty}&461\\
\h{tlot}&442\\
\h{politics}&379\\
\h{gop}&363\\
\h{healthcare}&275\\
\h{p2}&249\\
\h{sgp}&239\\
\h{nobel}&217\\
\h{tea}&183
\end{tabular}
}
\subfigure[\h{apple}]{
\begin{tabular}{lr}
$h_j$&$C(h,h_j)$\\
\hline
\h{iphone}&1336\\
\h{mac}&815\\
\h{ipod}&233\\
\h{itues}&158\\
\h{google}&113\\
\h{imac}&105\\
\h{tech}&95\\
\h{microsoft}&95\\
\h{fail}&86\\
\h{snowleopard}&85
\end{tabular}
}
\subfigure[\h{windows}]{
\begin{tabular}{lr}
$h_j$&$C(h,h_j)$\\
\hline
\h{microsoft}&135\\
\h{mac}&70\\
\h{linux}&60\\
\h{vista}&57\\
\h{windows7}&51\\
\h{win7}&47\\
\h{software}&38\\
\h{xp}&37\\
\h{ubuntu}&34\\
\h{iphone}&30
\end{tabular}
}
\caption{Co-occurrence lists for three hashtags $h$.}
\label{dictexamples}
\end{table}

Three examples of the resulting dictionary entries can be seen in Table \ref{dictexamples}.
The results seem pretty reasonable, but the question is of course how to formally evaluate this dictionary. The least thing one would expect from such a dictionary is that the words in $D(h)$ are somehow related to $h$. My idea was that the intensity of a relation between two words can actually be measured by examining the path between them in the \emph{WordNet} \cite{wordnet} lexical database of hypernym/hyponym relations.

The main problem with this approach is, of course, that many hashtag names will not appear in the WordNet corpus, either because they are non-English words or because they are not real words at all. Still, if the dictionary provides strong relations for hashtags restricted on WordNet entries (resulting in a set $H_\text{WN}$ of 13,791 hashtags), this is good indication that it works in general.

Another issue is that WordNet does not deal with individual words (\emph{lemmas}), but with \emph{synsets}, which are simply groups of synonymous lemmas. The basic assumption of my evaluation is that two words are related as much as the most related pair of respective synsets is, or formally
\[ S(h_1,h_2):=\max_{\substack{\text{synset } s_1 \text{ s.t. } h_1\in s_1\\ \text{synset } s_2 \text{ s.t. } h_2\in s_2}} S(s_1,s_2),\]
where $S(\cdot,\cdot)$ denotes the similarity between two hashtags or two synsets, respectively.

For computing the actual similarity, two similarity measures were used:
\begin{enumerate}
\item
the \emph{path distance similarity}, which is defined as
\[ S_\text{path}(s_1,s_2):= \frac{1}{d(s_1,s_2)+1},\]
where $d(s_1,s_2)$ denotes the length between the two synsets $s_1$, $s_2$ in the taxonomy, and
\item
the \emph{Wu-Palmer distance} \cite{wupalmer}, which is defined as
\[ S_\text{WP}(s_1,s_2):=\frac{2d(s(s_1,s_2))}{d(s_1)+d(s_2)},\]
where $s(s_1,s_2)$ denotes the lowest common subsumer of $s_1$, $s_2$, and $d(s)$ the depth of synset $s$ in the taxonomy.
\end{enumerate}
Python functions for both measures are already implemented in the WordNet module of nltk \cite{nltk}.
The following calculations were performed:
\begin{enumerate}
\item\label{co:co}
For every hashtag $h$ in $H_\text{WN}$ and respective co-occurring hashtags $h_i\in D(h)\cap H_\text{WN}$ compute their similarities $S_{\text{path},\text{WP}}(h,h_i)$.
\item\label{co:twitter}
As a first baseline, for every hashtag $h$ in $H_\text{WN}$ and ten random hashtags $r_i\in H_\text{WN}$, compute their similarity $S_{\text{path},\text{WP}}(h,r_i)$.
\item\label{co:wordnet}
As a second baseline, for 10,000 lemmas $l$ in WordNet and ten random other lemmas $l_i$, compute their similarities $S_{\text{path},\text{WP}}(l,l_i)$.
\end{enumerate}
The first baseline acts as a baseline in Twitter, i.e. it discards co-occurrance information and, given a hashtag, considers similarities to arbitrary other hashtags. The second baseline is a general measure in WordNet, reflecting the average similarity among lemmas.

Taking averages of all respective similarities, the following results were obtained:

\begin{center}
\begin{tabular}{rl|c|c}
&& average $S_\text{path}$ & average $S_\text{WP}$ \\
\hline\hline
\ref{co:co}& Co-occurrences & 0.12 & 0.37 \\
\hline
\ref{co:twitter}& Baseline (Twitter) & 0.07 & 0.26 \\
\ref{co:wordnet}& Baseline (Wordnet) & 0.05 & 0.16
\end{tabular}
\end{center}

This shows a significantly higher similarity between co-occurring hashtags than between arbitrary pairs of hashtags and between random pairs of words. In particular, the WordNet path between co-occurring hashtags is only about half as long as between random hashtags or words.

\subsection{Clustered graph}
\label{clusteredgraph}

To visualize hashtags and their relations, I created a graph of the 1000 most frequent hashtags. Edges were created among the 600 pairs of hashtags with the highest co-occurrence counts, with a weight corresponding to this count. To find structures in the graph, it was partitioned into 20 parts using \emph{kmetis} \cite{metis,kmetis}. This graph partitioning basically minimizes the weight of edge cuts. The resulting graph was then ploted using the spring layout \cite{springlayout} in the \emph{NetworkX} library \cite{networkx}.

\begin{figure}
\centering
\subfigure[principal connected component]{
\label{fig:graph:all}
\includegraphics[width=0.5\textwidth]{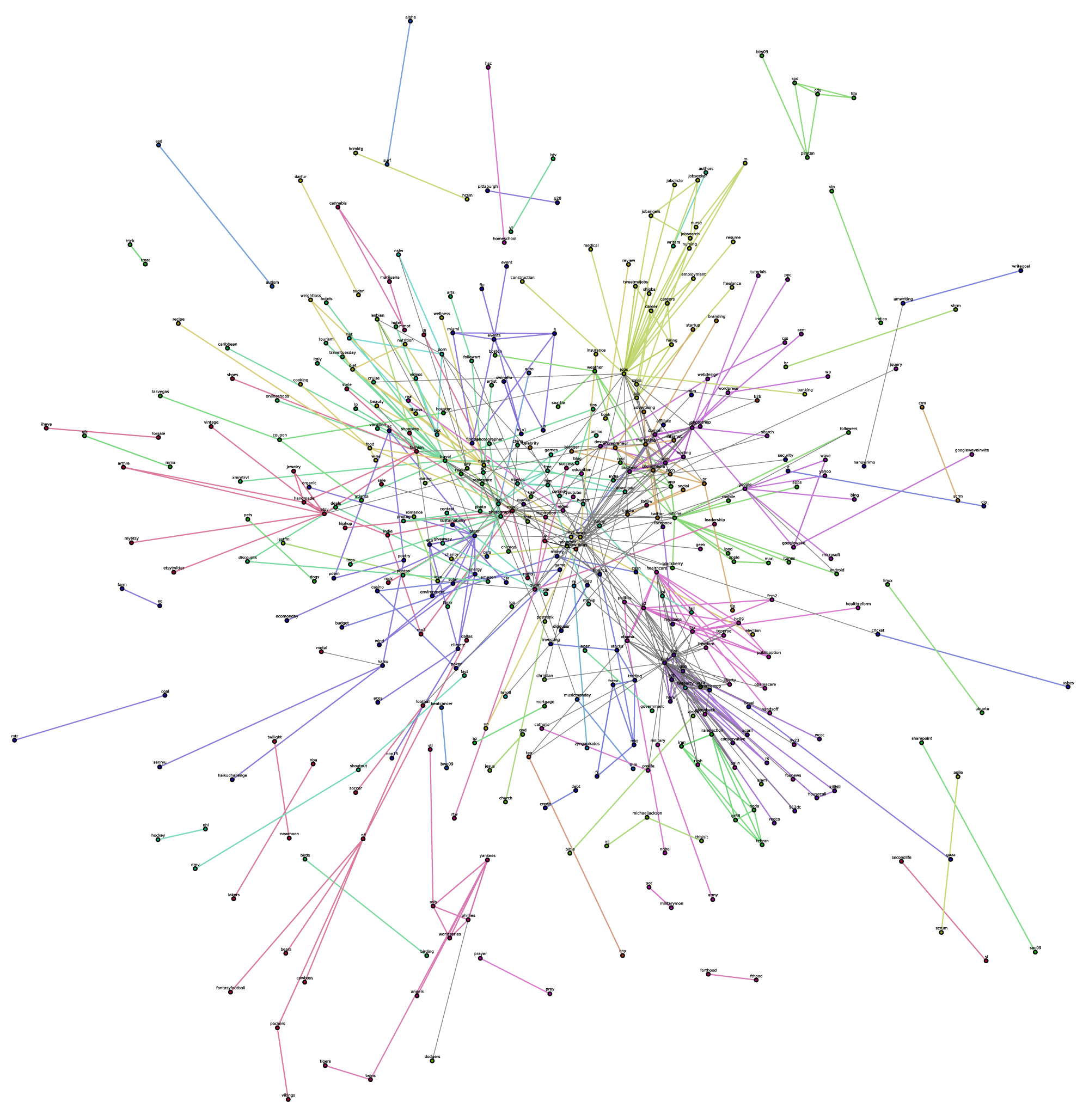}
}
\subfigure[German election cluster in the upper right corner, featuring political parties]{
\label{fig:graph:btw}
\includegraphics[width=0.4\textwidth,trim=0 5mm 0 0,clip=true]{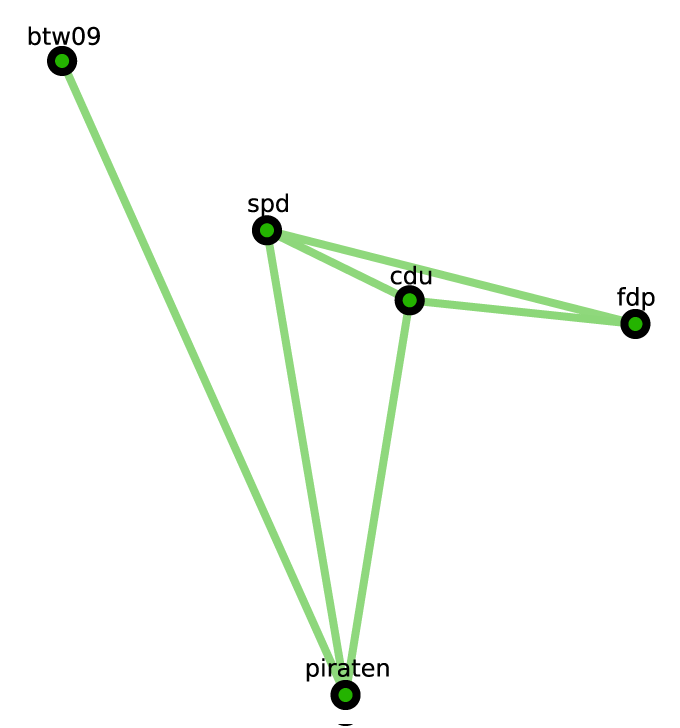}
}
\caption{Clustered graph of hashtags. Clusters are visualized by different colors, both for nodes and edges. Edges connecting nodes in different clusters are colored gray.}
\label{fig:graph}
\end{figure}

The result can be seen in Figure \ref{fig:graph:all}. It exhibits some really interesting relations among hashtags, and the clustering seems to catch many actual topic fields. For instance, there is a relatively clear cluster of hashtags related to U.S. politics, another one for jobs, and one for the German election in 2009; see Figure \ref{fig:graph:btw}.

\section{Classification of hashtags}
\label{classification}

\subsection{Classes}

Aside from retrieving related hashtags, it might be of particular interest to classify hashtags. This is related to named entity recognition: In most cases, hashtags represent named entities, apart from emotions, e.g. \h{fail}, or general categories, e.g. \h{photography}. Put it this way, the \emph{recognition} of (the most relevant) named entities in tweets is trivial---they are usually represented by hashtags.

What remains is the \emph{classification} of these named entities and other hashtags. As a first approach to this goal, I took the common named entity classes \emph{organization}, \emph{geolocation}, and \emph{person}, and added the classes \emph{event}, as event recognition might be of particular interest on Twitter, and \emph{category} which basically contains all other hashtags that do not fit into any other class, like emotions, fields of interest, and even products. See Table \ref{classes} for some examples of hashtags for each category.

\begin{table}
\centering
\begin{tabular}{l|l|l|l|l}
\textbf{Geolocation}&\textbf{Person}&\textbf{Organization}&\textbf{Event}&\textbf{Category}\\
\hline
\h{europe} & \h{obama} & \h{google} & \h{easter} & \h{photography} \\
\h{scandinavia} & \h{gandhi} & \h{pinkfloyd} & \h{christmas} & \h{fotografie} \\
\h{uk} & \h{coelho} & \h{yankees} & \h{election} & \h{politics} \\
\h{sverige} & \h{madonna} & \h{greenpeace} & \h{btw} & \h{math} \\
\h{california} & \h{gates} & \h{uno} & \h{btw09} & \h{rock} \\
\h{bigapple} & \h{mj} & \h{nokia} & \h{duell09} & \h{fun} \\
\h{graz} & \h{schröder} & \h{nestle} & \h{tsunami} & \h{fail} \\
\h{göteborg} & \h{berlusconi} & \h{bp} & \h{sziget} & \h{ipod}
\end{tabular}
\caption{Hashtag classes with examples.}
\label{classes}
\end{table}

\subsection{Machine-learning}
\label{machinelearning}

The basic approach was to use a maximum entropy (MaxEnt) classifier \cite[pages 235--241]{slp} to soft-classify the use of a hashtag in tweets it occurs in, and then classify the hashtag as the average of all these classifications. MaxEnt was chosen mainly because it performs very well in \cite{ner} on the one hand, and is readily implemented in nltk on the other hand. Employing \emph{SciPy} \cite{scipy}, the advanced limited memory Broyden–Fletcher–Goldfarb–Shanno (L-BFGS) algorithm \cite{lbfgs} can be used to solve the underlying convex optimization problem.

Given a hashtag $h$ in the text of a tweet, the following (binary-encoded) classification features were used:
\begin{enumerate}
\item
the words in a window of size 5 around $h$, excluding the hashtag itself,
\item
the shape feature \cite[pages 764--765]{slp} of each of these words (see Table \ref{shapefeatures}),
\item
the part-of-speech tags of these words (see section \ref{pos}),
\item
geographical background knowledge for these words (see section \ref{geo}),
\item
the shape feature of $h$ itself, without the leading hash (\verb!#!),
\item
an indicator whether $h$ is the first word (token) in the tweet,
\item
a position indicating in which fifth of the tweet (with respect to word indices) $h$ is,
\item
an indicator whether all the words following $h$ in the tweet are hashtags,
\item
the five most co-occurring hashtags of $h$, and
\item
geographical background knowledge for them.
\end{enumerate}

\begin{table}
\centering
\begin{tabular}{lll}
Shape&Example\\
\hline
@-reply&\verb!@jan!\\
hashtag&\verb!#example!\\
link&\verb!http://example.com!\\
number&\verb!123!\\
symbol & \verb!$!\\
ends in 1 digit & \verb!A1!\\
ends in 2 digits & \verb!btw09!\\
ends in 3 digits & \verb!N900!\\
ends in 4 digits & \verb!y2000!\\
contains digits & \verb!SLK300a!\\
all lower-case & \verb!lower!\\
all upper-case & \verb!UPPER!\\
first character capitalized, rest not & \verb!Sverige!\\
mixed capitalization & \verb!eBay!
\end{tabular}
\caption{Shape features used in the classifier.}
\label{shapefeatures}
\end{table}

\subsection{Part-of-speech tagging}
\label{pos}

As part-of-speech tagger, I chose \emph{HunPos} \cite{hunpos}, which is an open source reimplementation of the popular TnT tagger \cite{tnt}. It has an easy-to-use interface included in nltk and trained models for English text available for download. The latter might not be completely adequate for Twitter text, but the results are reasonable, so no further work was invested in that direction, so far. Here are two part-of-speech-tagged sentences from the introductory example:
\begin{quote}
\small
\begin{verbatim}
Will/MD make/VB more/JJR 4/CD u/NN. Live/JJ performances/NNS in/IN
#bolder/NN area/NN will/MD be/VB on/IN http://www.saxy.us/JJ.
\end{verbatim}
\end{quote}


\subsection{Geographical background knowledge}
\label{geo}

To ease the recognition of geolocations, features indicating whether a word is a city, region, or country name, respectively, are provided, complemented by a feature indicating whether the word is either of these. The set of geospatial names was acquired from \emph{Geonames} \cite{geonames} and includes names of 
219,833 cities having a population of at least 1000 inhabitants, 29,615 administrative regions, and 497 countries, each including alternate names in languages other than English.

\subsection{Evaluation}

For training the classifier and its evaluation, a set of 41 organization, 40 geolocation, 26 person, 16 event, and 57 category hashtags was classified by hand, yielding a total of 180 hashtags as ``gold standard.'' To reduce computational effort, not \emph{all} tweets containing a certain hashtag are used for training resp. classification, but only 100 random tweets.

The classifier was evaluated using 5-fold cross-validation. I chose five subsamples just because of execution time, as processing ten subsamples would have taken too long. They are selected randomly, consisting of 36 human-labeled hashtags each. So there are five evaluation phases, where in each the MaxEnt classifier is trained using $4\times 36=144$ hashtags and 100 random tweets each. Then the remaining 36 hashtags are classified by computing the average of the classifications of, again, 100 random tweets each.

The resulting confusion matrix can be seen in Table \ref{confusionmatrix}. Classification of geolocations and categories is ``okay'' (with a lot of room for improvement, still), while classification of events does not work at all, unfortunately.


\begin{table}
\centering
\renewcommand{\arraystretch}{1.15}
\[ \begin{array}{cr|rrrrr}
&& \multicolumn{5}{c}{\textbf{Hand-labeled class}}\\
& &\text{Category}&\text{Event}&\text{Geolocation}&\text{Organisation}&\text{Person}\\
\hline
\multirow{5}{5mm}{\begin{sideways}
{\textbf{Classification}}\end{sideways}} &
\text{Category} & 34 & 5 & 7 & 14 & 6 \\
&\text{Event} & 3 & 1 & 0 & 2 & 2 \\
&\text{Geolocation} & 6 & 1 & 23 & 6 & 2 \\
&\text{Organization} & 11 & 5 & 7 & 16 & 7 \\
&\text{Person} & 3 & 4 & 3 & 3 & 9 \\
\hline
&\text{Precision } P & 0.52&0.13&0.61&0.35&0.41\\
&\text{Recall } R & 0.60&0.06&0.58&0.39&0.35\\
&F_1&0.55&0.08&0.59&0.37&0.38
\end{array} \]
\caption{Confusion matrix of the hashtag classifier according to 5-fold cross-validation.}
\label{confusionmatrix}
\end{table}

\section{Web application}
\label{webapp}

To browse hashtags, their co-occurrence dictionaries, and classifications, I created the Web application\emph{TwitterExplorer}. It is built using the \emph{Django} Web framework\footnote{\url{http://www.djangoproject.com/}}. The JavaScript libraries \emph{Prototype}\footnote{\url{http://www.prototypejs.org/}} and \emph{RGraph}\footnote{\url{http://www.rgraph.net/}} are used for interface design.
To reduce page sizes, classification details for individual tweets are loaded using AJAX requests.
As of this writing, the application is publicly available at
\url{http://twex.poeschko.com}.

\begin{figure}
\centering
\includegraphics[width=0.8\textwidth]{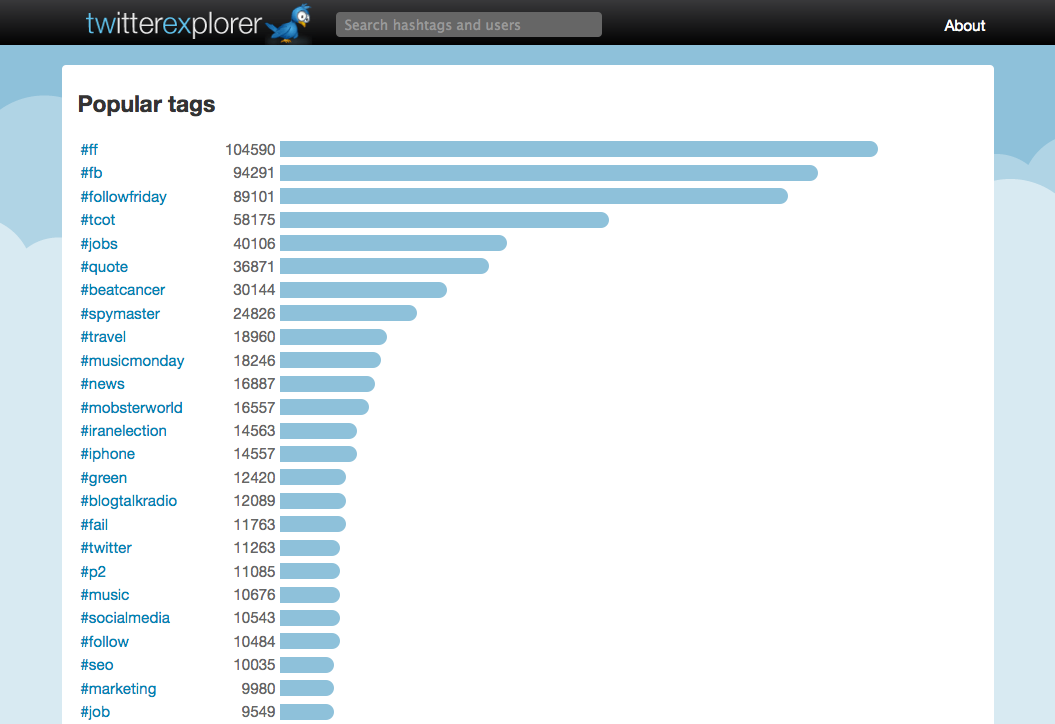}
\caption{Start screen of the TwitterExplorer Web application, showing a list of the most common hashtags.}
\label{startpage}
\end{figure}

\begin{figure}
\centering
\includegraphics[width=0.8\textwidth]{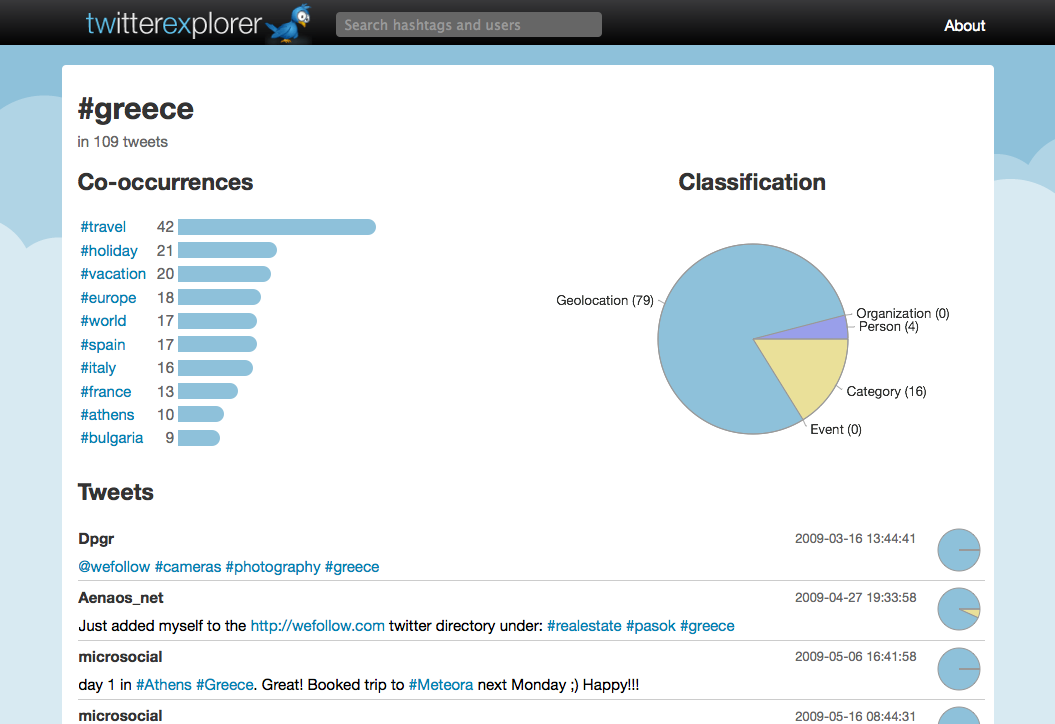}
\caption{Page for an individual hashtag (\h{greece}) in TwitterExplorer, showing the co-occurrence dictionary, overall classification of the hashtag, and the corresponding 100 tweets used for classification, including the respective classification. Detailed information about the classification can be displayed by clicking on the small pie charts to the right of each tweet.}
\label{hashtagpage}
\end{figure}

\section{Future work}

The are a number of ways in which this work could be improved or extended:
\begin{itemize}
\item
So far, only the \emph{precision} of the co-occurrence dictionary is evaluated, i.e. the question, ``how similar are the retrieved hashtags to the given hashtag.'' It would also be interesting to measure its \emph{recall}, i.e. ``how many of the similar hashtags are actually retrieved.''
\item
The clustered graph presented in section \ref{clusteredgraph} is only available as separate image file, so far. It would be nice to include it in the Web interface somehow, maybe complemented by other graph illustrations of hashtags and their ``surroundings.''
\item
The hashtag classifier certainly needs further improvement. In general, more training examples and a closer investigation of the feature selection might help. In order to enable event detection, features like the time distribution of tweets (e.g. its entropy) are certainly needed.
\item
The dataset also includes ``social'' information in the form of follower/following relationships. Maybe this can be employed in the classifier as well as in the ways hashtags can be browsed in TwitterExplorer. Moreover, the ``social relevance'' of Twitter hashtags will be subject to further research.
\item
Detecting hypernomy/hyponomy relations among hashtags would be a nice additional feature.
It might be done statistically or by machine learning.
\end{itemize}

\bibliography{literature}{}
\bibliographystyle{plain}

\end{document}